\documentclass[sigconf]{acmart}
\usepackage{enumitem}
\usepackage{multirow}
\usepackage{breakurl}

\hypersetup{
    breaklinks=true,   
    colorlinks=true,   
    pdfusetitle=true,  
}
\AtBeginDocument{%
  }

\copyrightyear{2024}
\acmYear{2024}
\setcopyright{rightsretained}
\acmConference[RecSys '24]{18th ACM Conference on Recommender Systems}{October 14--18, 2024}{Bari, Italy}
\acmBooktitle{18th ACM Conference on Recommender Systems (RecSys '24), October 14--18, 2024, Bari, Italy}
\acmDOI{10.1145/3640457.3688040}
\acmISBN{979-8-4007-0505-2/24/10}

\begin{document}

\title{Enhancing Performance and Scalability of Large-Scale Recommendation Systems with Jagged Flash Attention}

\author{Rengan Xu, Junjie Yang, Yifan Xu, Hong Li, Xing Liu, Devashish Shankar, Haoci Zhang\\
Meng Liu, Boyang Li, Yuxi Hu, Mingwei Tang, Zehua Zhang, Tunhou Zhang, Dai Li\\
Sijia Chen, Gian-Paolo Musumeci, Jiaqi Zhai, Bill Zhu, Hong Yan, Srihari Reddy}

\email{{renganxu, junjieyang, xuyifan, hongli, xingl, devashish, haocizhang}@meta.com
}
\email{{mengliu2019, boyangli, yuxihu, mingwt, zehua, tunhouzhang, daili1}@meta.com}
\email{{sijiac, gpmusumeci, jiaqi.zhai2, billzh, hong, sriharir}@meta.com
}

\affiliation{%
  \institution{Meta Platforms}
  \city{Menlo Park}
  \state{CA}
  \country{USA}
}
\renewcommand{\shortauthors}{Xu et al.}

\begin{abstract}
  The integration of hardware accelerators has significantly advanced the capabilities of modern recommendation systems, enabling the exploration of complex ranking paradigms previously deemed impractical. 
  However, the GPU-based computational costs present substantial challenges. 
  In this paper, we demonstrate our development of an efficiency-driven approach to explore these paradigms, moving beyond traditional reliance on native PyTorch modules. We address the specific challenges posed by ranking models' dependence on categorical features, which vary in length and complicate GPU utilization.
  We introduce Jagged Feature Interaction Kernels, a novel method designed to extract fine-grained insights from long categorical features through efficient handling of dynamically sized tensors.
  We further enhance the performance of attention mechanisms by integrating Jagged tensors with Flash Attention. Our novel Jagged Flash Attention achieves up to 9$\times$ speedup and 22$\times$ memory reduction compared to dense attention. Notably, it also outperforms dense flash attention, with up to 3$\times$ speedup and 53\% more memory efficiency. In production models, we observe 10\% QPS improvement and 18\% memory savings, enabling us to scale our recommendation systems with longer features and more complex architectures.
\end{abstract}





\maketitle

\section{Introduction}
Categorical features, such as user-clicked items within the last month, are heavily relied upon by ranking models~\citep{naumov2019deep,zhang2022dhen,zhang2024wukong,zhai2024actions,tang2024async}. Unlike dense (float) features, which maintain a fixed size across training samples, the length of categorical feature values can vary among different training samples. Padding has traditionally been used to standardize the sizes of these categorical features across various training samples within a batch~\citep{ptnested21,tfragged22}. However, while padding can be sufficient in some cases, it has inherent drawbacks. These are particularly noticeable with long length categorical inputs, a common input format for large, complex models trained on GPUs. Padding can introduce significant overhead, leading to increased memory usage, computational demands, and communication overhead. This not only affects the model's efficiency but also hampers scalability, particularly in environments with limited resources.

In this paper, we present our efforts in designing and adopting an efficiency-driven approach in the exploration of computationally expensive ranking paradigms. This shift marks a departure from the conventional approach of relying solely on the combination of native PyTorch modules to achieve algorithmic logic. In the rest of the paper, we will discuss the challenges, the methodologies and the lessons learned from our journey. By sharing our experiences, we aim to contribute to the collective knowledge base of the RecSys community, empowering fellow researchers and practitioners to navigate similar challenges in their pursuit of innovation.
\section{Methodology}
\begin{figure}[b]
    \vspace{-2em}
    \begin{center}
        \includegraphics[width=0.4\textwidth]{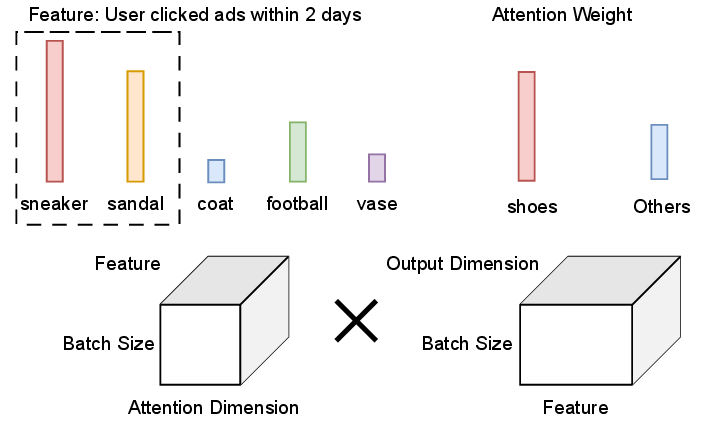}
        \vspace{-0.5em}
        \caption{Jagged Feature Interaction Kernel.}
        \label{fig:jagged_feat_interact}        
    \end{center}
\end{figure}

\begin{table*}[t]
\caption{Triton Kernels Benchmark Result for Jagged Tensor Operators. $B$: Batch size. $sum\_B$ is the jagged dimension with the total sequence length across samples in a batch. $Bi$: Sequence length for sample $i$. $D$: Embedding dimension. $T$: Hyperparameter. PyTorch version: The implementation with native PyTorch operator which requires padding. Triton version: The implementation with custom Triton operator without padding.}
\vspace{-1em}
\begin{center}
\scalebox{0.9}{
\begin{tabular}{|l|l|l|l|l|l|}
\hline
\textbf{GPU Kernels for Jagged tensors} & \textbf{Description} & \textbf{Version}  & \textbf{Memory (MB)} & \textbf{FLOPs (M)} & \textbf{Latency (us)} \\ \hline
\multirow{2}{*}{jagged\_dense\_bmm} & $[sum\_B, D] \times [B, D, T] = [sum\_B, T]$ 
 & \textit{PyTorch} & 309 & 838.9 & 166.6 \\ 
 & & \textbf{Triton} & 109 (2.83$\times$)  & 422.5 (1.98$\times$) & 99.3 (1.67$\times$) \\
 \hline
\multirow{2}{*}{jagged\_jagged\_bmm} & $[sum\_B, D]\times [sum\_B,T]=[B,D,T]$ &\textit{PyTorch} & 206 & 838.9 & 199.4 \\
 & & \textbf{Triton} & 104 (1.98$\times$) & 422.5 (1.98$\times$) & 79.1 (2.52$\times$)\\ \hline
\multirow{2}{*}{jagged\_softmax} & $sum\_(softmax([Bi, D]))$ & \textit{PyTorch} & 12.5 & 4.9 & 24.9 \\
 & & \textbf{Triton} & 6.3 (1.98$\times$) & 2.5 (1.98$\times$) & 18 (1.38$\times$) \\
\hline
\multirow{2}{*}{jagged\_jagged\_bmm\_jagged\_out} & $[sum\_B, D] \times [sum\_B, D] = [sum\_(Bi*Bi)]$ &\textit{PyTorch} & 1680 & 20971 & 671 \\
 & & \textbf{Triton} & 540 (3.11$\times$) & 7200 (2.91$\times$) & 293 (2.29$\times$) \\ \hline
\multirow{2}{*}{array\_jagged\_bmm\_jagged\_out} & $[sum\_(Bi*Bi)] \times [sum\_B, D] = [sum\_B, D]$ &\textit{PyTorch} & 4330 & 20971 & 755 \\
 & & \textbf{Triton} & 1990 (2.18$\times$) & 7200 (2.91$\times$) & 585 (1.29$\times$) \\ \hline
\multirow{2}{*}{jagged2\_softmax} & $sum\_(softmax([Bi*Bi]))$ &\textit{PyTorch} & 1530 & 122.9 & 2162  \\
 & & \textbf{Triton} & 520 (2.94$\times$) & 42.2 (2.91$\times$) & 707 (3.06$\times$) \\ \hline
\end{tabular}
}
\end{center}
\vspace{-1em}
\label{tab:kernel}
\end{table*}

We propose Jagged Feature Interaction Kernel, an innovative method tailored for extracting fine-grained insights from long categorical features. 
Figure \ref{fig:jagged_feat_interact} demonstrates an overview of our proposed kernel. 
By focusing on the interactions between feature values and targeting items, Jagged Feature Interaction prioritizes the most relevant feature values, assigning them higher weights. The features are represented with Jagged tensor from TorchRec~\citep{ivchenko2022torchrec}. The jagged tensor efficiently stores variable-length features from multiple samples in a compact and contiguous manner within memory without the need for padding. We achieve this using two tensors: one for holding all feature values collectively and another offset tensor that determines the sample boundaries for each feature segment. 

\subsection{Jagged Flash Attention}
The flash attention \citep{dao2022flashattention,dao2023flashattention} is the state-of-the-art algorithm for accelerating the standard attention. Its core idea is to fuse separate attention operations into a single kernel, minimizing data movement between GPU shared memory and global memory, and maximizing computations within the fast shared memory. Similar to the classic matrix multiplication optimization, flash attention employs tiling optimization to perform two GEMM operations and one softmax block by block. However, applying softmax independently to each block poses a challenge, as it requires the sum of exponentials in the denominator, which depends on information from later blocks. To overcome this challenge, it leverages the online softmax algorithm \cite{milakov2018online}, adjusting results for later blocks as new information becomes available.
The flash attention optimization could be applied in both dense and jagged attention. To achieve the best performance and maximize the memory saving, we have combined both jagged tensor and flash attention into jagged flash attention optimization. 

\subsection{Triton Kernels for Jagged Tensor}
We built customized Triton kernels~\cite{tillet2019triton} 
for both forward and backward computations for Jagged tensor operations. Triton is the programming paradigm based on blocked algorithms which can facilitate the construction of high-performance compute kernels for neural networks and allow compilers to aggressively optimize programs for data locality and parallelism. Specifically, we build 
\begin{itemize} 
  \item Jagged Tensor (sparse) multiply Jagged Tensor (sparse)
  \item Jagged Tensor (sparse) multiply Dense Tensor (dense)
  \item Softmax for Jagged Tensor
  \item MLPs for Jagged Tensor
  \item Elementwise operations for Jagged Tensor
  \item Jagged flash attention
  \item Conversions between Jagged Tensor and Dense Tensor
\end{itemize}

\section{Experiments and Conclusion}
\begin{figure}[t]
    \centering
    \includegraphics[width=0.45\textwidth]{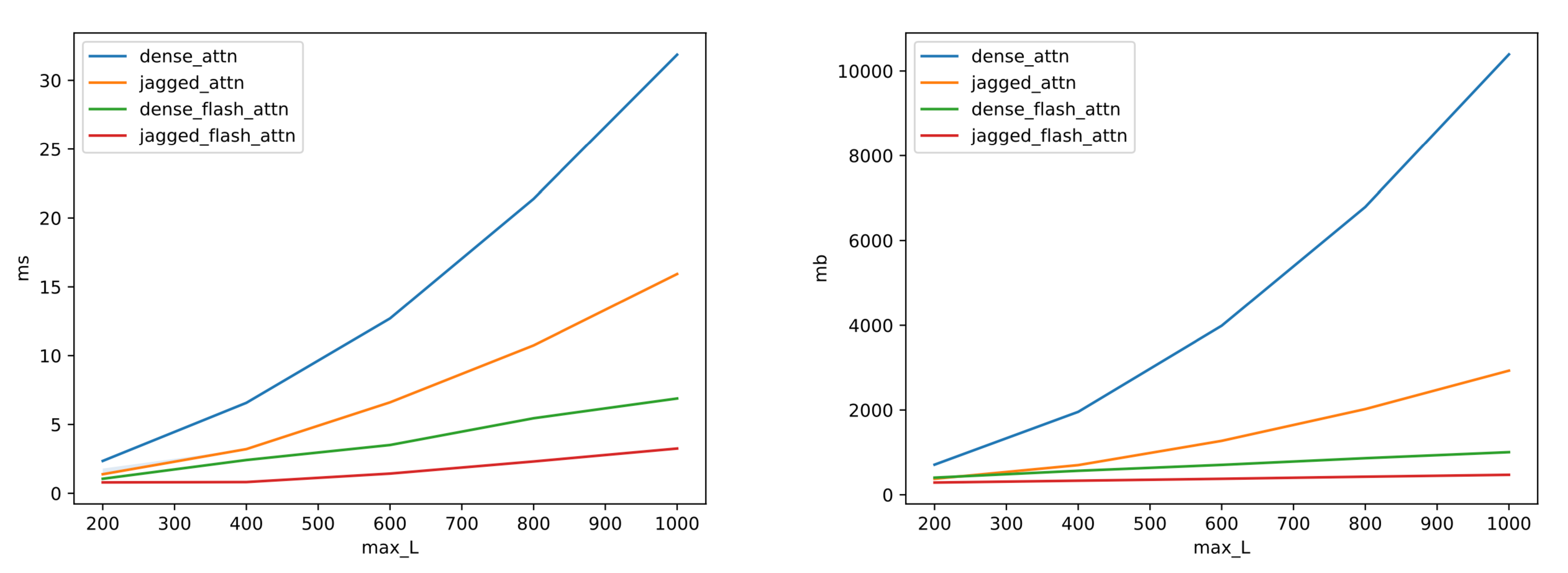}
    \vspace{-1em}
    \caption{Latency \& memory benchmark results for attention. $max\_L$: max sequence length}
    \label{fig:attn_bench}
    \vspace{-2em}
\end{figure}

Table~\ref{tab:kernel} shows the performance comparison between the custom Triton and the native PyTorch implementations for selective kernels. We demonstrate the relative improvement of Triton over PyTorch in parenthesis. It can be seen that the jagged operators reduce the FLOPs and memory usage significantly and outperform the dense version accordingly. 

We also compared different attention implementations with BF16 in Figure~\ref{fig:attn_bench}. The jagged attention mechanism offers significant speedup and memory efficiency improvements over dense attention. Specifically, jagged attention achieves up to 2$\times$ speedup compared to dense attention, while jagged flash attention further improves this to up to 9$\times$. Even when compared to dense flash attention, jagged flash attention still offers up to 3$\times$ speedup.
In terms of memory usage, jagged attention is up to 3.5$\times$ more efficient than dense attention, while jagged flash attention reduces memory usage by up to 22$\times$. Notably, the memory usage for both dense and jagged flash attention increases linearly rather than quadratically, with jagged flash attention being up to 53\% more memory efficient.
These improvements have practical implications for end-to-end model training, where we have observed approximately 10\% QPS improvement and 18\% memory savings for production models. This enables us to scale our recommendation systems further, accommodating longer features and more complex model architectures.

\begin{acks}
This work is a collective endeavor of many individuals, and the invaluable contributions from the following people (listed in alphabetical order): Adnan Akhundov, Wei Chen, Yang Chen, Lu Fang, Carl Hu, Yuzheng Huang, Mu-Chu Lee, Bert Maher, Andrey Malevich, Sarunya Pumma, Ketan Singh, Adele Sun, Xiaodong Wang, Wei Wei, Xinfeng Xie, Jackie Xu, Chunzhi Yang, and Mengchi Zhang. We would also like to thank Sandeep Pandey, Santanu Kolay, Ajit Mathew, Damien Sereni, Ian Barber, Colin Taylor, Adnan Aziz for their leadership support. 
\end{acks}

\bibliographystyle{ACM-Reference-Format}
\bibliography{acmart}

\end{document}